# Preliminary investigation into how limb choice affects kinesthetic perception

Mohit Singhala,[1] *Member, IEEE*, Amy Chi, Maria Coleman and Jeremy D. Brown, *Member, IEEE*

*Abstract*— We have a limited understanding of how we integrate haptic information in real-time from our upper limbs to perform complex bimanual tasks, an ability that humans routinely employ to perform tasks of varying levels of difficulty. In order to understand how information from both limbs is used to create a unified percept, it is important to study both the limbs separately first. Prevalent theories highlighting the role of central nervous system (CNS) in accounting for internal body dynamics seem to suggest that both upper limbs should be equally sensitive to external stimuli. However, there is empirical proof demonstrating a perceptual difference in our upper limbs for tasks like shape discrimination, prompting the need to study effects of limb choice on kinesthetic perception. In this manuscript, we start evaluating Just Noticeable Difference (JND) for stiffness for both forearms separately. Early results validate the need for a more thorough investigation of limb choice on kinesthetic perception.

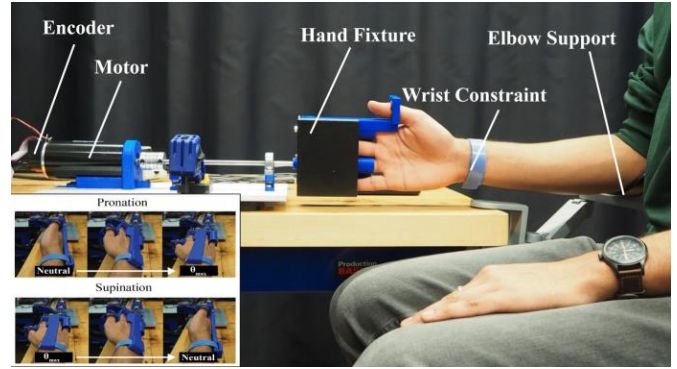

Fig. 1. Experimental setup with a motor, encoder, custom hand fixture, wrist constraint and elbow support; standard exploratory motion (bottom left)

## I. INTRODUCTION

Humans routinely use bimanual actions to perform tasks of varying difficulty. Surgeons, in particular, perform complex bimanual operations in tasks with low margins of error with a high degree of success, guided in part by their haptic perception as they use tools to interact with tissue. With the rise in use of Human-in-the-loop telerobotics systems (HiLTS) like the da Vinci, there is a need for higher quality haptic feedback to improve the transparency of such systems [1]. Understanding how humans process kinesthetic information from the environment in bimanual operations will be important to aid these efforts.

At present, however, there is a lack of consensus in the literature on how the human brain integrates information from both hands to generate a unified percept [2]. While some studies have shown empirical evidence supporting a difference in perception thresholds for our upper limbs [3], such differences are often overlooked in haptic studies that involve both limbs, like contralateral psychophysical tasks [4]. The prevalent theory that our central nervous system (CNS) can account for internal body dynamics while exploring external stimuli also seems to suggest that there should be no difference in perception based on the limb being used for exploration [5], [6], [7].

An important step towards understanding how we integrate information in bimanual tasks is to understand the similarities and dissimilarities between the two upper limbs during unimanual exploration. To this effect, we have started evaluating the Just Noticeable Difference (JND) for stiffness for both forearms using a single degree of freedom kinesthetic feedback device. Based on the results shown by Squeri et. al. [3], who establish a difference in cutaneous perception between both hands, we also expect to see a difference in the JNDs between both forearms of each participant.

## II. METHODS

### A. Participants

We recruited n=10 individuals (8 male, 2 female, age = 22.5$\pm$1.4 years) to distinguish virtual torsional springs with different spring constants. The participants were compensated at a rate of $10/hour. All participants were consented according to a protocol approved by the Johns Hopkins School of Medicine Institutional Review Board (Study# IRB00148746).

### B. Experimental Setup

The experimental apparatus uses a custom direct drive 1-DoF rotary kinesthetic haptic device (Fig. 1). The device features a Maxon RE50 370356 (200 Watt) motor equipped with a 3-channel Maxon Encoder HEDL #110518 (500 CPT) encoder and is driven by a Quanser AMPAQ-L4 Linear current amplifier. Data acquisition and control will be provided through a Quanser QPIDe PCI data acquisition card with a MATLAB/Simulink and Quarc real-time software interface run at a frequency of 1KHz. A custom 3D printed hand fixture, attached via a rotary shaft, serves as the primary mode of interaction for the participant. The fixture is designed to enable a unique alternating finger grip whose purpose is to limit flexion and extension of the whole hand. Wrist flexion

*This material is based upon work supported by the National Science Foundation under NSF Grant# 1657245.

[1]Mohit Singhala, Amy Chi, Maria Coleman and Jeremy D. Brown are with the Department of Mechanical Engineering, Johns Hopkins University, Baltimore, MD, USA. `mohit.singhala@jhu.edu`

and extension is minimized with the use of a hook-and-strap, enabling a uniform grip across all participants. The elbow is placed on a height-adjustable support to align the forearm axis of rotation with the device's rotational axis and to limit radial and ulnar deviation.

*C. Procedure*

JND estimates for stiffness discrimination were obtained separately for explorations with the right and left forearms of the participant. Participants performed a standard two-alternative forced choice (2AFC) same-different task in a single sitting consisting of two sessions separated by a five-minute break. Each session consisted of multiple trials where participants were presented with a reference virtual torsional spring *(k = 2 mNm/deg)* and a test virtual torsional spring. The order of presentation of the springs was randomized for each trial and the order of exploration for both forearms was randomized for each participant. Participants explored the two springs one at a time and reported if they felt the same or different. Each exploration required pronation of the forearm from the normal position to the maximum angular displacement position and supination back to normal (see Figure 1). The stiffness of the test spring was subsequently changed based on the staircase algorithm described in Section II-D. All participants filled out the Edinburgh Handedness survey during the break between the two JND sessions.

*D. Staircase*

A weighted 1 up/3 down staircase algorithm was used to determine the MDTs. Based on values suggested by Garcia-Perez [8], the up step-size was set at 10% of the reference stimuli and the ratio of down step-size and up step-size was 0.7393 for a proportion correct target of 83.15%. The staircase was initialized at 1.5 times the value of the reference stimuli. The spring torque was rendered according to the following relationship:

$$\tau = s \cdot k \cdot \theta \quad (1)$$

where $k$ is the spring constant, $s \in [1, 2]$ is a scaling factor whose value was determined by the staircase algorithm, and $\theta$ was the angle of the participant's hand in the device. For the reference spring s = 1.

The staircase was terminated after ten reversals and the average of the last eight reversals is reported as the JND. A 1 up/1 down approach was followed until the first reversal to prevent presenting too many springs above the participant's threshold as recommended [9], [10]. If the participant missed the target position by more than 2.5 degrees, the trial was deemed unsuccessful. These trials were repeated before moving ahead on the staircase. Participants were informed to treat repetitions as a fresh trial and the springs were repeated as a pair in random order.

## III. RESULTS

Based on the Edinburgh handedness survey scores ($E_s$), seven participants demonstrated strong right hand dominance ($E_s \geq 50$), two demonstrated strong left hand dominance ($E_s \leq 0$), and one participant showed mild right hand dominance ($0 < E_s < 50$). The JNDs for both forearms of all participants are plotted in Fig. 2. Seven out of the ten participants demonstrated lower JNDs when exploring with their right forearm, including both of the left-handed participants. The participant with mild right hand dominance performed better with their left forearm.

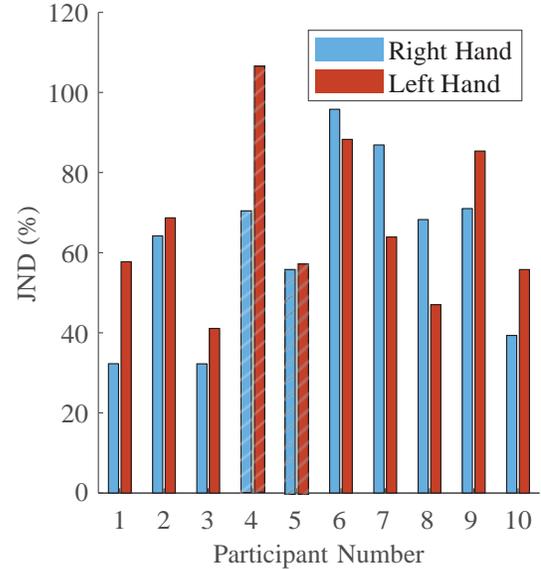

Fig. 2. JNDs for both forearms for each participant. Hatched patterns represent data from Left hand dominant participants

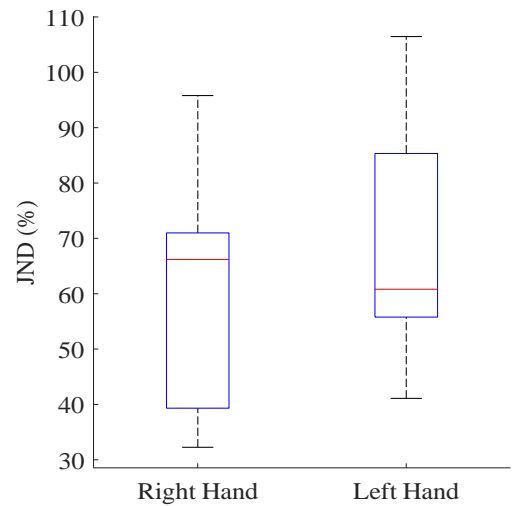

Fig. 3. JNDs for the right and left forearm of each participant

We used MATLAB 2018b and IBM SPSS 25 for all of our statistical analyses. Normality in each of the JND distributions we looked at (right and left hand, dominant and non-dominant hand) was satisfied based on Shapiro-Wilk test results. Figure 3 shows the JND of the participants' right forearms and left forearms. In comparing right-forearm and left-forearm performance, we found no significant difference between JNDs using a two-sample t-test.

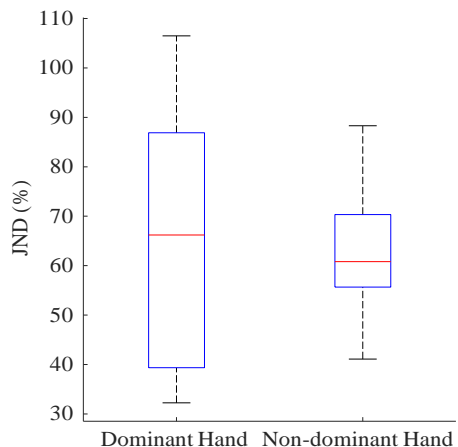

Fig. 4. JNDs for the dominant and non-dominant hand of each participant

Figure 4 shows a comparison of the participants' dominant and non-dominant forearms. Again, we did not find a significant difference in JND between the two categories. As such, we then looked at each individuals' best and worst forearms in terms of JND. The difference between best and worst forearms is depicted in Figure 5. We then conducted a one-sample t-test to determine if this difference between individuals best and worst performances is significant; the t-test revealed that at an $a$ level of 0.05, the difference is indeed significant with a t-score of 4.67 and a p-value of .0012.

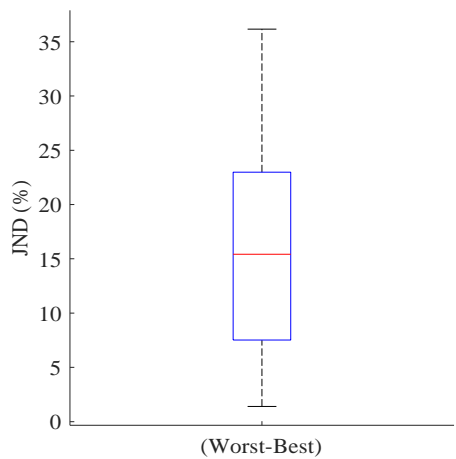

Fig. 5. Absolute difference in JND values for the two forearms

## IV. DISCUSSION

We first note that the thresholds in our experiment are considerably higher than those reported in literature for stiffness. We believe that this is likely a result of the fact that we allowed participants to explore each spring only once. The physical action of rotating the forearm to explore a torsion spring is also different from the standard practice of using hands or fingers to manipulate linear springs [11].

This lack of significant results for comparison of Right and Left forearms is expected, as we are not aware of any literature that would indicate a statistically significant advantage for the right forearm over the left forearm or vice versa. Although we see that neither handedness nor hand dominance appear to be playing a significant role in perception, with no significant differences found between right-forearm & left-forearm JNDs and dominant & non-dominant JNDs, and both left-handed participants performing better with their right forearms, we did find that on an individual level there is a significant difference in perception between each individuals' two forearms. It has been well-established that we can distinguish the sensory consequences of our own actions from externally produced sensory stimuli [7], giving us the ability to monitor and recognize our own, self-generated limb movements. However, our results suggest that this may not hold true for our experiment, despite the external stimuli and the exploration methodology being identical for both forearms. We believe that investigating this phenomenon further will be essential to the design of haptic feedback systems for HiLTS. However, the lack of results when comparing dominant and non-dominant forearms prompts the need for further investigation of these concepts; we expected that the individual differences could be explained by the dominance of the better limb, but currently, we don't have any significant evidence to show this to be true. Overall, the results seem to validate the need for a more thorough investigation into the potential effects of our exploration strategies on our haptic perception.

## ACKNOWLEDGMENT

This material is based upon work supported by the National Science Foundation under NSF Grant 1657245.